\def\year{2020}\relax
\documentclass[letterpaper]{article} % DO NOT CHANGE THIS
\usepackage{aaai20}  % DO NOT CHANGE THIS
\usepackage{times}  % DO NOT CHANGE THIS
\usepackage{helvet} % DO NOT CHANGE THIS
\usepackage{courier}  % DO NOT CHANGE THIS
\usepackage[hyphens]{url}  % DO NOT CHANGE THIS
\usepackage{graphicx} % DO NOT CHANGE THIS
\usepackage{graphicx}  % DO NOT CHANGE THIS
\usepackage{array}
\usepackage{multirow}
\usepackage{booktabs}
\usepackage{subcaption}
\usepackage{graphbox}
\usepackage{amsmath}
\usepackage{amssymb}
\usepackage{arydshln}

\newcommand{\fakepar}[1]{\noindent\textbf{#1}}
\newcommand{\etal}{et al.}
\newcommand{\figcontent}[2]{
  \includegraphics[page=#1,width=.95\columnwidth]{{5440_figures_preview}.pdf}
}
\newcommand{\figcontentwide}[2]{
  \includegraphics[page=#1,width=1.95\columnwidth]{{5440_figures_preview}.pdf}
}

\newcommand{\hp}{\hat{p}_i^t}
\newcommand{\hf}{\hat{f}_i}

\newcommand{\figintro}{
\begin{figure}[t]
  \centering
  \figcontent{1}{\figintrocontent}
  \caption{The face reenactment results of our self-supervised disentangling method.
  We use the target frames (in the first row) to puppeteer source images (in the first column) into desired poses.
  Such animation can be performed at both the same and different identities.
  }
  \label{fig:intro}
\end{figure}
}

\newcommand{\figpipeline}{
\begin{figure*}[t]
  \centering
  \figcontentwide{2}{\figpipelinecontent}
  \caption{The training pipeline of our framework.
  Given a talking face sequence, DAE-GAN encodes the identity and pose features respectively, which are subsequently fed to a conditional generator for synthesizing photo-realistic and  pose-alike images.
  Specifically, the face embedder encodes multiple frames into one embedded face $\hat{f}_i$, while the pose embedder extract identity-independent pose code $\hp$ from each frame.
  The genertor network maps input face  $\hat{f}_{i}$ and extracted pose vector $\hp$ into the output $\tilde{x}_{i}^t$ which preserve the identity of source image and act as the driving image.
  In the training stage, we only optimize our model with frames from the same video sequence; during the testing, the images fed to the face embedder and the pose embedder can come from different identities.
  The generator is able to flawlessly synthesize the photo-realistic faces with source identity and target poses.
  Notice that the sampling operation implements a bilinear sampling kernel on the input.
  }
  \label{fig:pipeline}
\end{figure*}
}

\newcommand{\figpose}{
\begin{figure}[t]
  \centering
  \figcontent{3}{\figposecontent}
  \caption{Multi-scale pose embedder encodes a frame $x^t_i$ into the pose code $\hp$ and
  decodes it to a displacement field and two residual maps.
  The warping field samples from the embedded face to reconstruct the input at low resolution.
  The output then is added with two high-resolution residual maps to restore the original image.}
  \label{fig:pose_embedder}
\end{figure}
}

\newcommand{\figface}{
\begin{figure}[t]
  \centering
  \figcontent{4}{\figfacecontent}
  \caption{Multi-frame face embedder. It encodes multiple faces and predicts a displacement field $\mathcal{T}$ and an attention map $\mathcal{A}$ for each frame. Then these outputs are fused to generate an embedded face $\hat{f}_i$.}
  \label{fig:face_embedder}
\end{figure}
}

\newcommand{\tabcomparison}{
\begin{table}[t]
  \centering
  \resizebox{.95\columnwidth}{!}{\begin{tabular}{lccc}
    \toprule
    Methods ($T$)  & SSIM$\uparrow$ & FID$\downarrow$ & USER$\downarrow$ \\
    \midrule
    \multicolumn{4}{c}{\textbf{VoxCeleb1}} \\
    \midrule
    Pix2PixHD (1)$^*$
                   & 0.56           & 42.7            & 0.82 \\
    Pix2PixHD (8)$^*$
                   & 0.64           & 35.1            & 0.79 \\
    Pix2PixHD (32)$^*$
                   & 0.70           & \textbf{24.0}   & 0.71 \\
    X2Face (1)     & 0.68           & 45.8            & 0.82 \\
    X2Face (8)     & 0.73           & 51.5            & 0.83 \\
    X2Face (32)    & \textbf{0.75}  & 56.5            & 0.85 \\
    Zakharov \etal (1)$^*$
                   & 0.67           & 43.0            & 0.62 \\
    Zakharov \etal (8)$^*$
                   & 0.71           & 38.0            & 0.62 \\
    Zakharov \etal (32)$^*$
                   & 0.74           & 29.5            & \textbf{0.61} \\
    \midrule
    Ours (vanilla w/ $\cal{L}_\text{REC}$)
                   & 0.65           & 60.8            & -    \\
    + smoothness loss $\cal{L}_\text{S}$
                   & 0.68           & 56.9            & -    \\
    + multi-frame  & 0.72           & 47.1            & -    \\
    + multi-scale  & 0.73           & 46.8            & -    \\
    + adversarial learning (w/ $\cal{L}_\text{GAN}+\cal{L}_\text{R}$)
                   & 0.73           & 26.3            & -    \\
    + feature matching loss $\cal{L}_\text{FM}$
                   & 0.73           & 24.8            & \textbf{0.61} \\
    \midrule
    \multicolumn{4}{c}{\textbf{RaFD}} \\
    \midrule
    FaceSwapNet$^*$& 0.71           & 12.3            & -   \\
    Ours           & 0.73           & 13.8            & -   \\
    \bottomrule
  \end{tabular}}
  \caption{Quantitative comparisons of methods on VoxCeleb1-test and RaFD dataset. The number in the parentheses are the size $T$ of finetuning set. Note that we do not finetune our models during testing. Symbol $^*$ denotes that landmark information is used in this method.}
  \label{tab:comparison}
\end{table}
}

\newcommand{\figcomparison}{
\begin{figure*}[t]
  \centering
  \figcontentwide{5}{\figcomparisoncontent}
  \caption{Comparisons on the VoxCeleb1 test dataset. All of the source and target images are not shown in the training process. To demonstrate the generalization performance, we conduct both intra-identity and inter-identity experiments. Note that the results of Zakharov \etal in the intra-identity part are from the original paper, and those in the inter-identity part are generated by our re-implemented model.}
  \label{fig:comparison}
\end{figure*}
}

\newcommand{\figdisentangle}{
\begin{figure}[t]
  \centering
  \figcontent{6}{\figdisentanglecontent}
  \caption{Self-supervised identity-pose disentangling on the VoxCeleb1 test set.
  Our model learns independent representations for identity and pose through a reconstruction training process.
  The embedded face and the warping field are visualized in (c) and (d). The former tends to be a canonical appearance while the latter learns a global transformation.}
  \label{fig:disentangle}
\end{figure}
}

\newcommand{\figquery}{
\begin{figure}[t]
  \centering
  \figcontent{7}{\figquerycontent}
  \caption{Image retrieval using the embedded pose vector.
  This embedding is proved to have encoded the pose-related information
  since the retrieved images have similar poses and emotions but different identity with the query image.}
  \label{fig:query}
\end{figure}
}

\newcommand{\figsamplesrafd}{
\begin{figure}[t]
  \centering
  \figcontent{8}{\figsamplesrafdcontent}
  \caption{Results on the RaFD datasets. For each source face, driving poses are taken from different identities.}
  \label{fig:samplesrafd}
\end{figure}
}

\title{Realistic Face Reenactment via Self-Supervised Disentangling of Identity and Pose}
\author{
\Large \textbf{
Xianfang Zeng,$^*$
Yusu Pan,\thanks{Equal contribution. Names are in alphabetical order.}
Mengmeng Wang,
Jiangning Zhang,
Yong Liu\thanks{Corresponding author}
} \\
Institute of Cyber-Systems and Control, Zhejiang University, China \\
\{zzlongjuanfeng, corenel, mengmengwang, 186368\}@zju.edu.cn, yongliu@iipc.zju.edu.cn \\
}
\begin{document}

\maketitle

\begin{abstract}
Recent works have shown how realistic talking face images can be obtained under the supervision of geometry guidance, e.g., facial landmark or boundary.
To alleviate the demand for manual annotations, in this paper, we propose a novel self-supervised hybrid model (DAE-GAN) that learns how to reenact face naturally given large amounts of unlabeled videos.
Our approach combines two deforming autoencoders with the latest advances in the conditional generation.
On the one hand, we adopt the deforming autoencoder to disentangle identity and pose representations.
A strong prior in talking face videos is that each frame can be encoded as two parts: one for video-specific identity and the other for various poses.
Inspired by that, we utilize a multi-frame deforming autoencoder to learn a pose-invariant embedded face for each video.
Meanwhile, a multi-scale deforming autoencoder is proposed to extract pose-related information for each frame.
On the other hand, the conditional generator allows for enhancing fine details and overall reality. It leverages the disentangled features to generate photo-realistic and pose-alike face images.
We evaluate our model on VoxCeleb1 and RaFD dataset. Experiment results demonstrate the superior quality of reenacted images and the flexibility of transferring facial movements between identities.
\end{abstract}

\section{Introduction}
\label{sec:introduction}

Face reenactment aims at transferring facial movements and expressions from one driving video to another source face/video.
Such ability holds promise to an abundance of applications like face editing, movie making, video conferencing and augmented reality.
This task is known to be challenging for two main reasons.
Firstly, under the uncontrolled condition, the appearance of the monocular face is determined by several coupled factors such as identity, pose, expression and reflection, etc.
This intrinsically entangled characteristic makes it hard to transfer a particular attribute between faces.
For example, identity-related features like face outline in the source image are usually changed during the process of mimicking facial movements in a driving video~\cite{CycleGAN2017}.
The second aspect is the low tolerance of the human visual system for minor mistakes in generated images.
Some hardly captured defects like local blur or unnatural expression will significantly reduce the reality of synthesized pictures.

\figintro

Several methods have been proposed to overcome the challenges, including the classical parametric models and data-driven learning frameworks.
Parametric 3D face models~\cite{blanz1999morphable} can provide a fully-controllable representation for manipulating a predefined face.
Nevertheless, those models usually are unable to capture all subtle movements of the human face without delicate designs.
On the other hand, benefiting from recent remarkable advances in image generation~\cite{goodfellow2014generative,park2019SPADE}, data-driven frameworks~\cite{wu2018reenactgan,Zakharov2019FewShotAL} have performed extremely realistic face reenactment with geometry guidance, e.g., facial landmark/boundary~\cite{guo2019pfld}.
However, the premise of such approaches is the numerous manual annotations of facial landmark, which is expensive and time-consuming.

To alleviate the demand for adequate and accurate annotations, we propose a hybrid model (DAE-GAN) for reenacting talking faces in a self-supervised manner.
Only with the assumption of the availability of talking face videos, our model can learn to puppeteer a source face given a driving video.
Our approach combines deforming autoencoder~\cite{shu2018deforming} and generative adversarial networks~\cite{Wang2017HighResolutionIS} to achieve both representation disentanglement and high-quality image synthesis.
In short, two deforming autoencoders are utilized to disentangle identity and pose features, which subsequently is fed to a conditional generator to synthesize photo-realistic and pose-alike face image.

Specifically, we observe a strong prior in talking face videos that all frames share fundamental features such as identity while each frame keeps transferable representations like pose, expression, etc.
Inspired by that, a multi-frame deforming autoencoder is used to fuse different frames and estimates an embedded face for one video.
The embedded face, therefore, tend to be in a frontalised view, which is a pose-free representation and contains only identity features.
Meanwhile, a multi-scale warping network is proposed to capture the global deformation in each frame.
We constrain this network to reconstruct different frames from the embedded face. Hence, it is forced to extract the pose-related information of each frame.
Subsequently, to synthesize photo-realistic and pose-alike face image, the embedded face is fed to the generator as input while the pose vector as a condition. This adversarially-trained deep convolutional network provides high-frequency details and overall reality.

In experiments, quantitative and qualitative comparisons are conducted on VoxCeleb1 and RaFD dataset~\cite{Nagrani2017VoxCelebAL,zhang2019faceswapnet}.
In addition, we show several applications of our methods, including image retrieval across identities as well as intra/inter-identity face puppeteering.
The results demonstrate the superior quality of reenacted images and the flexibility of transferring facial movements between identities.

In summary, our contributions are two folds:
1) We propose a self-supervised framework to naturally reenact talking faces through watching large amounts of unlabeled videos.
Experimental results show that DAE-GAN outperforms the state-of-the-art self-supervised methods and is comparable to the approaches with geometry guidance.
2) The proposed multi-frame/scale deforming autoencoders can disentangle the identity and pose representations. It indicates potential applications in image retrieval.
% section Introduction (end)

\section{Related Work}
\label{sec:related_work}

\fakepar{Parametric modeling for face manipulation.}
One of the most classic parametric face models is 3DMM \cite{blanz1999morphable}.
Later works build upon the fitting of 3DMM by introducing high-level details~\cite{saito2017photorealistic}, or learning 3DMM parameters directly from RGB data~\cite{tewari2017mofa}.
Face puppeteering can be performed by fitting the face model and then manipulate the estimated parameters.
For instance, given a driving and source video sequence, Face2Face \cite{thies2016face2face} models both the driving and source face via a 3DMM or 3D mesh.
The estimated 3D face model is used to transform the expression of the source face to match that of the driving face.
Recently, \citeauthor{averbuch2017bringing} propose a 2D warping method to reenact a facial performance given only a single target image.
The target image is animated through 2D warps that imitate the facial transformations in the driving video.
In general, the parametric model can provide control over the facial parameters and allows for explicit manipulation of the facial attributes.
Nevertheless, a 3DMM approach is limited by the components of the corresponding morphable model, which may not model the full range of expressions/deformations.
The pre-defined 3D model, therefore, can hardly capture all subtle movements of the human face.
% subsection Face Reenactment (end)

\fakepar{Learning based methods for face reenactment.}
Benefiting from large-scale face database collections~\cite{langner2010presentation,Nagrani2017VoxCelebAL} and reliable landmark detection techniques~\cite{guo2019pfld}, numerous impressive face reenactment methods are proposed in a way of direct synthesis of video frames.
In early efforts, \citeauthor{xu2017face} utilize CycleGAN~\cite{CycleGAN2017} to transfer face expressions between two identities.
However, in this model, the identity-related features like face outline are usually changed during the process of face reenactment.
In contrast to CycleGAN-based methods, ReenactGAN~\cite{wu2018reenactgan} maps all faces into a boundary latent space and then decodes it to each specific person.
Introducing the boundary space improves facial action consistency and the robustness for extreme poses.
More recently, FaceSwapNet~\cite{zhang2019faceswapnet} is presented to extend ReenactGAN for solving a more flexible many-to-many face reenactment problem.
Moreover, \citeauthor{Zakharov2019FewShotAL} present a system for creating talking head models from a handful of photographs.
Such an approach is able to learn highly realistic and personalized talking head models in a few-shot manner after a meta-learning.

However, all the above methods except CycleGAN-based ones perform the face reenactment task with the assumption of the availability of facial landmark/boundary.
In contrast to that, our method learns the geometry guidance from videos in a self-supervised way.
Through the disentanglement of identity and pose representations, our model is able to reenact face naturally between different identities.
% section Representation Disentanglement (end)

\fakepar{Self-supervised representation disentanglement.}
Self-supervised learning adopts supervisory signals that are inferred from the structure of the data itself~\cite{zhang2017split,Wiles18a}.
For face analysis, \citeauthor{shu2018deforming} introduce Deforming Autoencoders that disentangles shape from the appearance in a self-supervised manner.
\cite{Wiles2018X2FaceAN} present X2Face to implicitly learn a face representation from an extensive collection of video data.
Very recently, TCAE~\cite{li2019self} is presented to change the facial actions respectively and head poses of the source face to those of the target face. The trained model, therefore, can disentangle the facial action related movements and the head motion related ones.

\figpipeline

\section{Proposed Method}
\label{sec:methods}
We aim at transferring facial movements from a driving video onto a source face in a self-supervised manner.
Figure~\ref{fig:pipeline} illustrates the training framework of DAE-GAN given a talking face sequence.
The full architecture involves two embedders to decouple identity and pose features, a pair of generator-discriminator to synthesize photo-realistic reenacted face.
Below we present details of the four parts:

\begin{itemize}
  \item The face embedder $F(x_i^1,...,x_i^k) \rightarrow \hat{f}_i$ takes multiple frames from one video and maps them into an embedded face.
  We denote with $x_i$ the $i$-th video sequence, $x_i^t$ its $t$-th frame and $\hat{f}_i$ the embedded face for $i$-th video.
  The embedded face is expected to contain only identity features invariant to the pose and expression in each frame.
  \item The pose embedder $P(x_i^t; \hat{f}_i) \rightarrow \hat{x}_i^t$ takes as inputs the embedded face and frames with different pose.
  The pose embedder is designed to reconstruct different frames from the embedded face. Hence, it is forced to extract pose-related information from each frame.
  \item The conditional generator $G(\hat{p}_i^t; \hat{f}_i) \rightarrow \tilde{x}_i^t$ leverages the extracted pose vector and the embedded face to generate a pose-alike image. Here, we denote $\hat{p}_i^t$ with the extracted pose vector of $x_i^t$. The generator is trained to minimize the distance between $\tilde{x}_i^t$ and its ground truth $x_i^t$.
  \item The discriminator $D(\tilde{x}_i^t; x_i^t)$ takes the synthesized image and the corresponding video frame to distinguish detailed differences between them.
  This discriminator provides an adversarial learning objective together with the generator to synthesize photo-realistic images.
\end{itemize}

\subsection{Disentanglement of Identity and Pose}
\label{sub:disentanglement}
Disentanglement of identity and pose aims at learning independent representations for them.
We argue that the face generation can be interpreted as a combination of two processes: a synthesis of the deformation-free face template, followed by global deformation which involves pose information.
a multi-frame deforming autoencoder is utilized as the face embedder, which takes multiple faces to generate an embedded face.
Meanwhile, a multi-scale deforming autoencoder is proposed as the pose embedder to extract pose information for each frame.

As can be seen in Figure~\ref{fig:face_embedder}, the face embedder predicts a displacement field $\mathcal{T} \in \mathbb{R}^{W \times H \times 2}$ and an attention map $\mathcal{A} \in \mathbb{R}^{W \times H}$ for each input where $W$ and $H$ are the width and height of the images.
$\mathcal{T}_{u,v} = (\delta{u}, \delta{v})$ is the flow vector for pixel $(u,v)$ in source image. Namely, the pixel $(u, v)$ is moved to the location $(u+\delta{u}, v+\delta{v})$ in target image.
The displacement field give us a spatial transformed face $\mathcal{T}_k(x_i^k)$ through a bilinear sampling ~\cite{jaderberg2015spatial} from source image.
The attention map provides relative importance when fusing multiple generated faces into an embedded face. This process can be formulated as
\begin{equation}
  \begin{aligned}
    \hat{f}_i = \sum_{k=1}^K \mathcal{A}_k^{\prime} \odot \mathcal{T}_k(x_i^k) \text{,}
  \end{aligned}
  \label{eq:embedded_face}
\end{equation}
where $\mathcal{A}_k^{\prime} = {\mathcal{A}_k} / (\sum_{k=1}^K \mathcal{A}_k)$ means the normalized weight.

The pose embedder takes a frame as input and learns to transform pixels from the embedded face to reconstruct the input.
As shown on Figure~\ref{fig:pose_embedder}, it has an encoder-decoder architecture with multi-scale outputs, which contain a displacement field $\mathcal{T}^{-1} \in \mathbb{R}^{(W / 4) \times (H / 4) \times 2}$ and two residual maps  $\mathcal{R}^L \in \mathbb{R}^{(W / 2) \times (H / 2) \times 3}$, $\mathcal{R}^H \in \mathbb{R}^{W \times H \times 3}$, respectively.
Instead of making the pose embedder directly predict a warping field of $W \times H$ resolution, we consider decomposing the reconstruction into two parts: a global deformation learning and a fine details enhancement.
We account for the global deformation by a warping field $\mathcal{T}^{-1}$ at low resolution. It is encouraged to ignore some local warps and to pay more attention to the global tendency.
The residual maps $\mathcal{R}^L$, $\mathcal{R}^H$ aim at learning the elaborate differences between generated image and its ground truth.
The reconstructed frame $\hat{x}_i^t$ can be interpreted as that $\mathcal{T}^{-1}$ samples the main content from the embedded face while $\mathcal{R}^L$ and $\mathcal{R}^H$ enhance image details. It is formulated as
\begin{equation}
  \begin{aligned}
    \hat{x}_i^t = \mathcal{R}^H_t + U\{ \mathcal{R}^L_t + U[\mathcal{T}_t^{-1}(\hat{f}_i)] \} \text{,}
  \end{aligned}
  \label{eq:reconstruct_face}
\end{equation}
where $U$ is an upsampling operator for the summation of images in different resolutions.
In order to sample correctly from the embedded face into frames with a different pose, the pose vector is encouraged to encode pose/expression/other factors.

\figface

During the training process of representation disentanglement, the parameters of the face embedder and the pose embedder are optimized to minimize the objective including the reconstruction term and the smoothness term.
The reconstruction term $\mathcal{L}_\text{REC}$ measures the L1 distance between the ground truth image $x_i^t$ and the reconstruction frame $\hat{x}_i^t$. That gives:
\begin{equation}
  \begin{aligned}
  \mathcal{L}_\text{REC}\left(x_i^t, \hat{x}_i^t \right)=\left\|x_i^t - \hat{x}_i^t\right\|_{1} \text{.}
  \end{aligned}
  \label{eq:reconstruction_loss}
\end{equation}

The smoothness term penalizes quickly-changing in displacement field to avoid self-crossing in the local deformation. In particular, it measures the total variation of warping fields horizontally and vertically, denoted as
\begin{equation}
  \begin{aligned}
    \mathcal{L}_\text{S}(\mathcal{T}, \mathcal{T}^{-1}) =
    & \mathbb{E}( \left\| \nabla_u\mathcal{T} \right\|_{1} + \left\| \nabla_v\mathcal{T} \right\|_{1} ) + \\
    & \mathbb{E}( \left\| \nabla_u\mathcal{T}^{-1} \right\|_{1} + \left\| \nabla_v\mathcal{T}^{-1} \right\|_{1} ) \text{.}
  \end{aligned}
  \label{eq:smoothness_loss}
\end{equation}

The full objective for two embedders can be denoted as:
\begin{equation}
  \begin{aligned}
    \mathcal{L}\left(F, P\right)=\mathcal{L}_\text{REC}+\lambda_\text{S}\mathcal{L}_\text{S} \text{,}
  \end{aligned}
  \label{eq:disentanglement}
\end{equation}
where the hyper-parameter $\lambda_\text{S}$ is set $1$ in our experiment.
% subsection Disentanglement of Identity and Pose (end)

\subsection{Adversarial Learning}
\label{sub:adversaraial}
The goal of our proposed adversarially-trained submodule is enhancing high-frequency details and overall reality of the generated images.
This submodule consists of a conditional image generator $G$ and a discriminator $D$.
The generator $G$ takes a source image $\hf$ as input and uses conditional normalization layers to fuse the extracted pose code $\hp$ into the output.
As shown in the Figure~\ref{fig:pipeline}, our generator $G$ is built on the architecture proposed by \citeauthor{Wang2017HighResolutionIS}, which has been proven successful for image translation.
Different from the original Pix2PixHD setting, we replace all the convolution blocks with residual ones \cite{He2015DeepRL} and self-attention module \cite{Zhang2018SelfAttentionGA}.
Besides, we use the adaptive instance normalization (AdaIN) layer \cite{Huang2017ArbitraryST} in the residual blocks of the middle part $G_m$ and the back-end decoder $G_b$.
Each MLP block consists of one shared linear layer and two corresponding linear layers for mean and standard deviation outputs.

\figpose

We train this network by solving a minimax problem:
\begin{equation}
\min_{D}\max_{G} \mathcal{L}_{\text{GAN}}(D,G) +
\lambda_{\text{R}} \mathcal{L}_{\text{R}}(G) + \lambda_{\text{FM}} \mathcal{L}_{\text{FM}}(G) \text{,}
\label{eqn:adversarial_loss}
\end{equation}
where $\mathcal{L}_{\text{GAN}}$, $\mathcal{L}_{\text{R}}$, and $\mathcal{L}_{\text{FM}}$ are the GAN loss \cite{Lim2017GeometricG,Miyato2018SpectralNF}, the content reconstruction loss and the feature matching loss \cite{Salimans2016ImprovedTF,Gulrajani2017ImprovedTO} respectively.

We choose the hinge version of the adversarial loss in the alternative of the classic minimax loss for a more robust training:

\begin{equation}
  \begin{aligned}
    \mathcal{L}_{D} = & - \mathbb{E} [ \min (0, -1 + D(x^t_i)) ]   \\
                      & - \mathbb{E} [ \min (0, -1 - D(G(\hf, \hp)) ] \\
    \mathcal{L}_{G} = & - \mathbb{E} [ D(G(\hf, \hp)) ]
    \text{.}
  \end{aligned}
  \label{eqn:hinge_loss}
\end{equation}

The content reconstruction loss $\mathcal{L}_{\text{R}}$ encourages $G$ to generate an image identical to the driving image, while the feature matching loss $\mathcal{L}_{\text{FM}}$ regularizes the output of $G$ to match the ground truth in the representation space embedded by the discriminator $D$. Specifically, we measure $\mathcal{L}_{\text{R}}$ and  $\mathcal{L}_{\text{FM}}$ both by L1 norm with the same form as Equation~\ref{eq:reconstruction_loss}. In addition, the coefficient $\lambda_{\text{R}}$ and $\lambda_{\text{FM}}$ are both set to $1$.
% subsection Adversarial Learning (end)

\subsection{Training Strategy}
\label{sub:training_strategy}
In total, there are four networks in our framework.
To reduce training time and stabilize the training process, we utilize a two-stage strategy to train our full model.
In the early period of training, both the extracted embedded face and the pose vector are meaningless.
Hence, it is unnecessary to optimize the conditional generator until the training of representation disentanglement tends to be stable.
We experimentally determine to only optimize the two embedders in the first $30$ epochs.
Subsequently, the generator and discriminator are added to the optimizing group.

The first training stage aims at self-supervised disentangling of identity and pose representations.
Two embedding networks are trained through an image reconstruction process and the full objective is Equation~\ref{eq:disentanglement}.
This stage is sufficient to train the networks such that the pose embedder encodes expression and pose of the driving frame while the face embedder encodes identity information.

In the second training stage, we first decrease the learning rate of two embedders by a factor of $10$. Then, the generator and discriminator are optimized from the initial learning rate.
It is worth noting that they are optimized alternately instead of in an end-to-end manner.
Specifically, the four networks are optimized one by one. When one is under training, the parameters of the rest are frozen.
% subsection Training Strategy (end)
% section Proposed Methods (end)

\figcomparison

\section{Experiments}
\label{sec:experiments}

We first evaluate our model by performing an ablation study. We then provide a quantitative comparison against state-of-the-art methods. Finally, we show some qualitative results to demonstrate the performance of our model.

\fakepar{Implementation details.}
\label{par:implementation}
All experiments are conducted in a node with 2 NVIDIA RTX 2080Ti GPUs.
The learning rate is set to $1\times10^{-4}$, except for the discriminator, whose is $4\times10^{-4}$. We use the Adam \cite{Kingma2014AdamAM} optimizer with $\beta_1=0, \beta_2=0.9$
and decrease learning rate linearly.
% paragraph Implementation details (end)

\fakepar{Datasets.}
\label{par:datasets}
We conduct our ablation study and comparisons on VoxCeleb1 dataset \cite{Nagrani2017VoxCelebAL,Nagrani2018SeeingVA}. Face images in $256\times256$ resolution are extracted from the videos at $1$~fps. We train all the models on the training and validation set and report their results on the corresponding test set.
We also perform experiments on the RaFD dataset \cite{langner2010presentation}. Since this dataset contains only 8040 images with 67 identities, it cannot meet the diversity and scale requirement for our self-supervised method to train from scratch.
Therefore, we adapt our model trained on the VoxCeleb1 dataset and finetune it on the RaFD dataset for 20 epochs.

% paragraph Datasets (end)

\fakepar{Performance metrics.}
\label{par:metrics}
To quantify the quality of our results, we adopt the evaluation protocol from the previous work \cite{Zakharov2019FewShotAL}.
Specifically, we randomly select $50$ videos from the test set and $32$ hold-out frames from each video.
These frames are excluded from the fine-tuning process (if necessary) and used as driving images to be transformed from the remaining part in each video.

In terms of identity preservation, we use structured similarity (SSIM) \cite{Wang2004ImageQA} as a metric for the low-level similarity.
For the photo-realism, we use Fréchet Inception Distance (FID) \cite{Heusel2017GANsTB} to measure distribution distance between the real images and synthesized results. Moreover, we use the same settings for user study as \cite{Zakharov2019FewShotAL}.
% paragraph Performance metrics (end)

\fakepar{Baselines.}
\label{par:baselines}
We compare our method with four leading face generation and manipulation models: \citeauthor{Zakharov2019FewShotAL}, X2Face \cite{Wiles2018X2FaceAN}, Pix2PixHD \cite{Wang2017HighResolutionIS} and FaceSwapNet \cite{zhang2019faceswapnet}. \citeauthor{Zakharov2019FewShotAL} is the current start-of-the-art adversarial generative model for driving talking heads, which achieves few-shot learning by finetuning. Besides, X2Face takes a self-supervised approach without the requirement of annotations for training, while Pix2PixHD is a general GAN-based conditional image synthesis framework. Moreover, FaceSwapNet is a recent many-to-many face reenactment network. All baselines except X2Face are trained with extra landmarks.

For a fair comparison, we re-implement the unreleased model of \citeauthor{Zakharov2019FewShotAL} and use the officially provided model of X2Face. Specifically, for the model of \citeauthor{Zakharov2019FewShotAL}, we set the number of training frames as $1$ and finetune it by 40 epochs before inference. As the above models may not perform identically to the original paper, we use the results and images provided by the authors whenever available.
% paragraph Baselines (end)

\tabcomparison

\subsection{Ablation Study}
\label{sub:ablation}
We conduct an ablation study to verify the impact of each component in DAE-GAN.
The results are shown in Table~\ref{tab:comparison}.
The vanilla variation consists of two deforming autoencoders: a single-frame one for the face embedder and a single-scale one for the pose embedder.
The embedded face estimated by the single-frame autoencoder is usually incomplete, which further leads to visible local warps in the reconstruction image.
Our vanilla variation, therefore, gets the worst performance on both SSIM and FID.
The smoothness loss $\cal{L}_\text{S}$ is proposed to avoid self-crossing in the local deformation. There are some crossed artifacts in the generated images if we remove the smoothness item.
We then replace the single-frame autoencoder with a multi-frame autoencoder to test the effect of the multi-frame fusion mechanism.
As can be seen in Table~\ref{tab:comparison}, it brings significant performance improvements on SSIM and FID.
The main reason is that the fused embedded face is more complete than that in the single-frame model.
While there exists a global blur in the reconstructed face, it can learn the face structure from the input.
Furthermore, we can see that the blurring phenomenon is alleviated by adding the multi-scale mechanism on the pose embedder. Such sharper reconstructed images get better scores on two metrics.
Finally, a pair of generator and discriminator are utilized in our full methods. This adversarially-trained deep model is known to be excel at realistic image synthesis.
Its ability to capture high-frequency features can supplement numerous fine details on generated images.
Hence, we can see a significant improvement of photo-realism from the generated images and a noticeable decrease in FID score (lower is better).
The feature matching loss $\cal{L}_\text{FM}$ improves the performance slightly since it regularizes the synthesized images to match the real ones in the latent space of the discriminator $D$.
% subsection Ablation Study (end)

\subsection{Quantitative Comparisons}
\label{sub:quantitative}

As shown in Table~\ref{tab:comparison}, our full method outperforms all baselines which uses $1$ or $8$ images for finetuning, and still rivals baselines even when $T=32$.

We can find that X2Face (32) strikes the best SSIM with almost worst FID, while Pix2PixHD (32) performs best in FID but has a mid-stream SSIM.
We argue that this is due to the nature of different metrics. SSIM measures the structural similarity between two paired images, and mainly focuses on the low-level information of images, i.e., identity-preservation. On the other hand, FID measures the distribution distance between the two sets of images in the feature level, and chiefly focuses on the high-level information and more high-frequency details, i.e., photo-realism, which is just what adversarial methods are good at.
Both X2Face and Pix2PixHD are biased towards one aspect and neglect the other.
The architecture designs of \citeauthor{Zakharov2019FewShotAL} and our model take into account not only identity-preservation but also photo-realism so that these models can achieve extraordinary results on both FID and SSIM.

Compared with baselines, especially \citeauthor{Zakharov2019FewShotAL}, our method does not need any finetuning process and can match their tuned models with $T=32$. It indicates that our model not only has powerful embedders to separate the identity and pose, but also propose a stronger decoder to perform face reenactment work. Namely, our model can both maintain the source identity and produce more realistic images.

\figsamplesrafd

\figdisentangle

In Figure~\ref{fig:comparison}, we visualize the results of both intra-identity and inter-identity experiments. The results show that our model can successfully translate images to similar ones of target poses. The synthesized images are photo-realistic and resemble images from the target identity.
As for other baselines, the background of the image generated by X2Face produces a severe distortion in a complex environment, which is quite different from the original one. Moreover, in the inter-identity experiments, the generated faces are also distorted when given only one source image.
\citeauthor{Zakharov2019FewShotAL} performs quite well in the case of intra-identity transferring. However, if the source image and the target one are from different identities, its lack of landmark adaptation leads to a noticeable personality mismatch, especially when there exists significant differences in face poses.

Furthermore, we compare our model with FaceSwapNet on the RaFD dataset.
As shown in Table~\ref{tab:comparison} and Figure~\ref{fig:samplesrafd}, our model can achieve comparable results to FaceSwapNet.
The advantage of our model is that the whole process does not need extra annotations like landmark information.
% subsection Quantitative Comparisons (end)

\subsection{Qualitative Results}
\label{sub:qualitative}

\fakepar{Disentanglement.}
\label{par:distanglement}
We visualize the embedded faces $\hat{f}$ and displacement fields $\mathcal{T}$ of frames extracted from the same video. As shown in Figure~\ref{fig:disentangle}, our model learns to reconstruct the input images while automatically deriving the deformation-free face template and deformation-involving pose information. The generated embedded faces are almost the same, while the displacement fields indicate the different flows between each input frame and its corresponding embedded face. The results suggest that our model can correctly disentangle the identity and pose from a single image.
% paragraph Disentanglement (end)

\figquery

\fakepar{Image retrieval.}
\label{par:retrieval}
To demonstrate the power of our learned pose embedding, we conduct another experiment for retrieving images by facial attributes. This experiment is based on the intuition that if the two embedders indeed disentangle the identity and pose, the distance of extracted pose embeddings can be used to determine images with similar poses and facial attributes despite different people.
Firstly, we extract pose embeddings $\hat{p}_i$ of all images from VoxCeleb1 test set by the pre-trained pose embedder $P$. Given an image $x_q$, we encode its pose embedding $\hat{p}_q$, and then rank all test images $x_i$ according to the cosine similarity between $\hat{p}_q$ and $\hat{p}_i$.
The results are shown in the Figure~\ref{fig:query}, where the retrieved images have similar pose and emotions to those of the query images. These indicate that our pose embedding indeed encoded the facial attribute information and our embedders successfully disentangle the identity and pose.
% paragraph Image rettieval (end)
% subsection Qualitative Results (end)
% section Experiments (end)

\section{Conclusion}
\label{sec:conclusion}
We have proposed DAE-GAN, a self-supervised hybrid framework for reenacting talking faces, which is able to provide photo-realistic results and robustly disentangle the identity and pose representations. Crucially, our model needs neither manual annotations during training nor post-finetuning before inference. Even so, our model outperforms the state-of-the-art self-supervised methods and is comparable to those approaches with geometry guidance. Furthermore, we demonstrate its applications on image retrieval and facial expression transferring.
% section Conclusion (end)

\section{Acknowledgements}
\label{sec:acknowledgements}
This work is supported by the National Natural Science Foundation of China under Grant U1509210 and Key R\&D Program Project of Zhejiang Province (2019C01004).
% section Acknowledgements (end)

{
  % \small
  % \fontsize{9.0pt}{10.0pt}
  \bibliographystyle{aaai}
  \bibliography{5440.references}
}

\end{document}